\documentclass[letterpaper, 10 pt, conference]{ieeeconf}  
\pdfoutput=1

\IEEEoverridecommandlockouts                              

\usepackage{amsfonts}
\usepackage[cmex10]{amsmath}
\usepackage{array}
\usepackage{bbm}
\usepackage{blindtext}
\usepackage{booktabs}
\usepackage{cite}
\usepackage{cleveref}
\usepackage[pdftex]{graphicx}
\usepackage{multirow}
\usepackage{siunitx}
\usepackage{xcolor} 

\DeclareMathOperator*{\argmax}{arg\,max}

\crefname{figure}{Fig.}{Figs.}
\Crefname{figure}{Fig.}{Figs.}

\title{
Deep residual networks for automatic sleep stage classification of raw polysomnographic waveforms
}

\author{Alexander~N.~Olesen$^{\dagger,1,2}$,
        Poul~Jennum$^{3}$,
        Paul Peppard$^{4}$,
        Emmanuel Mignot$^{2}$,
        and~Helge~B.~D.~Sorensen$^{1}$,
\thanks{$^{\dagger}$Corresponding author: aneol@elektro.dtu.dk.}%
\thanks{$^{1}$Department of Electrical Engineering, Technical University of Denmark, Kgs. Lyngby, Denmark.}%
\thanks{$^{2}$Stanford Center for Sleep Sciences and Medicine, Stanford University, Palo Alto, CA, USA.}%
\thanks{$^{3}$Danish Center for Sleep Medicine, Department of Neurophysiology, Rigshospitalet, Glostrup, Denmark.}%
\thanks{$^{4}$University of Wisconsin School of Medicine and Public Health, Madison, WI, USA.}}

\begin{document}

\bstctlcite{IEEEexample:BSTcontrol}

\maketitle
\thispagestyle{empty}
\pagestyle{empty}

\begin{abstract}
We have developed an automatic sleep stage classification algorithm based on deep residual neural networks and raw polysomnogram signals. Briefly, the raw data is passed through 50 convolutional layers before subsequent classification into one of five sleep stages. Three model configurations were trained on 1850 polysomnogram recordings and subsequently tested on 230 independent recordings. Our best performing model yielded an accuracy of 84.1\% and a Cohen's kappa of 0.746, improving on previous reported results by other groups also using only raw polysomnogram data. Most errors were made on non-REM stage 1 and 3 decisions, errors likely resulting from the definition of these stages. 
Further testing on independent cohorts is needed to verify performance for clinical use.
\end{abstract}

\section{Introduction}

Sleep staging is the principal tool available to medical doctors in the analysis of sleep disorders. Natural human sleep consists of recurring cycles of three to four distinct phases, which are primarily characterized by changes in brain activity, eye movements, muscle activations and breathing. A polysomnogram (PSG) containing electroencephalography (EEG), electrooculography (EOG), electromyography (EMG) and other signals is collected during sleep, and subsequently
processed and analyzed by sleep technicians according to standards by the American Academy of Sleep Medicine (AASM). Each \SI{30}{\second} epoch of data is categorized into either  wakefulness (W), rapid eye movement (REM) sleep, or one of three stages of non-REM sleep (N1, N2, N3)~\cite{Berry2016}. However, this approach is prone to subjective interpretation of sleep staging rules, which have prompted extensive research in using various signal processing and machine learning approaches~\cite{Sen2014}.

Attempts at exploiting deep learning models for sleep staging have been proposed recently. One group used a transfer learning-approach to characterize sleep stages~\cite{Vilamala2017}, where 30 s epochs of Fpz-Cz EEG were subjected to multitaper spectral estimation (MTSE) in order to create spectral image representations~\cite{Thomson1982}, that ultimately were fed as input to a VGG-16 model stripped of the last layers~\cite{Simonyan2015}. This approach was cross-validated using a leave-one-out scheme on 20 subjects 
and yielded a bootstrapped accuracy of $86\%\pm2\%$.

Using MTSE for representing EEG data was also investigated in~\cite{Biswal2017}, where the authors compared various machine and deep learning models trained on either raw EEG waveforms, MTSE spectrograms, or 96 expert-defined features. They tested their best performing model on recordings from 1000 individual subjects and obtained an accuracy of \SI{85.76}{\percent} and a Cohen's kappa of 0.79 using a combination of expert-defined features and recurrent neural networks (RNN). On the same test set, they obtained accuracy/kappa values of \SI{77.31}{\percent} and 0.71 using a deep learning model trained on raw EEG waveforms.

However, it is still unclear whether manual feature extraction such as sleep spindle/K-complex detection, or data transformations, such as spectrograms or MTSE, are strictly necessary for efficient deep learning, and there is still room for improvement in the current state of the art for raw PSG analysis. 

We propose a novel method for automatic sleep staging combining state of the art deep learning networks with raw PSG data to accurately capture the complex relationships found in PSG data without resorting to data transformations and manual feature engineering.

\section{Data}
A database containing 2310 recordings extracted from the Wisconsin Sleep Cohort was used in this study. Specific acquisition details concerning the PSGs are described in~\cite{Young2008}. The entire set of PSG studies was randomly split into training (train), validation (eval), and testing (test) subgroups in an 8:1:1 ratio. Detailed demographic information as well as relevant PSG variables for all three subgroups are provided in~\cref{tab:wsc_demographics} including apnea-hypopnea index (AHI) and time spent in each sleep stage based on manual scoring. 

\begin{table}[tb]
    \footnotesize
    \renewcommand{\arraystretch}{1.3}
    \centering
    \caption{Extracted WSC cohort demographics for each subgroup. Significant \textit{p}-values are highlighted in bold.}
    \label{tab:wsc_demographics}
    \begin{tabular}{@{}lcccc@{}}
        \toprule
                                                & Train             & Eval              & Test              & \textit{p}-value \\ \midrule
        \textit{n} (male)                       & 1850 (1010)       & 230 (112)         & 230 (120)         & 0.210             \\
        Age (years)                             & $ 59.2 \pm 8.4 $  & $ 59.9 \pm 8.5 $  & $ 60.4 \pm 8.2 $  & 0.092             \\
        BMI (\si{\kilo\gram\per\metre\squared}) & $ 31.7 \pm 7.2 $  & $ 31.0 \pm 6.9 $  & $ 32.2 \pm 7.7 $  & 0.203             \\
        AHI (\si{\per\hour})                    & $ 12.6 \pm 15.6 $ & $ 11.5 \pm 14.9 $ & $ 12.4 \pm 16.2 $ & 0.600            \\ \midrule
        PSG dur. (\si{\hour})                & $ 7.4 \pm 0.8 $   & $ 7.4 \pm 0.7 $   & $ 7.4 \pm 0.8 $   & 0.947            \\ 
        W (\%)                                  & $ 18.5 \pm 11.3 $ & $ 17.2 \pm 11.1 $ & $ 19.6 \pm 11.8 $ & 0.071             \\
        N1 (\%)                                 & $ 8.2 \pm 4.5 $   & $ 8.8 \pm 5.6 $   & $ 8.9 \pm 5.1 $   & $\mathbf{0.038}$  \\
        N2 (\%)                                 & $ 54.2 \pm 10.3 $ & $ 54.0 \pm 10.9 $ & $ 52.4 \pm 11.0 $ & $\mathbf{0.048}$  \\
        N3 (\%)                                 & $ 5.8 \pm 6.4 $   & $ 6.4 \pm 7.0 $   & $ 6.0 \pm 7.0 $   & 0.433             \\
        REM (\%)                                & $ 13.3 \pm 5.9 $  & $ 13.7 \pm 5.8 $  & $ 13.2 \pm 5.7 $  & 0.635             \\ \bottomrule
    \end{tabular}
\end{table}

\section{Methods}

\subsection{Signal extraction and pre-processing}
Central and occipital EEG from right hemisphere, left and right EOG, and chin EMG channels were extracted from each PSG study. To accommodate different equipment setups used for recording studies, each channel was upsampled to \SI{200}{\hertz}. 
Following resampling, signals were filtered using zero-phase Butterworth filters with frequency ranges recommended by the AASM~\cite{Berry2016}. 
Since dynamic ranges vary considerably across channels, each signal was soft-normalized using the 5th and 95th quantiles, such that 
\begin{equation}
    \mathbf{x}_{\mathrm{norm}} = 2~\frac{\mathbf{x} - \mathrm{Q}_{0.05}(\mathbf{x})}{\mathrm{Q}_{0.95}(\mathbf{x}) - \mathrm{Q}_{0.05}(\mathbf{x})} - 1,
\end{equation}
where $\mathbf{x}_{\mathrm{norm}}$ denotes the normalized version of the signal $\mathbf{x}$, and $\mathrm{Q}_{0.05}(\mathbf{x})$ and $\mathrm{Q}_{0.95}(\mathbf{x})$ denotes the 5th and 95th percentile, respectively. Doubling and subtracting by one rescales $\mathrm{Q}_{0.05}(\mathbf{x})$ and $\mathrm{Q}_{0.95}(\mathbf{x})$ to $-1$ and $1$, respectively.

Finally, each signal was segmented into \SI{30}{\second} epochs corresponding to AASM criteria~\cite{Berry2016}, resulting in a tensor $\mathbf{X}$ with elements 
\begin{equation}
    (x_{nc\cdot t}) \in \mathbb{R}^{N\times C\times 1 \times T}, 
\end{equation} 
with $N=16$, $C=5$, and $T=6000$ being batch size, number of signals, and number of timesteps for one epoch, respectively.\footnote{The 1-dimensional convolution \texttt{tf.layers.conv1d} reshapes the input argument to subsequently call \texttt{tf.layers.conv2d} in TensorFlow. To reduce computational costs, we introduce a singleton dimension.} 


\subsection{Deep residual network model}
We applied a deep learning model inspired by the residual network models proposed in~\cite{He2016,He2016b}. These types of models employ residual skip connections between layers in order to maintain a proper gradient backpropagation through the network. This feature allows for extremely deep network structures, and a specific variant of this model with 152 layers came in 1st place in the ILSVRC '15 image classification competition~\cite{He2016}.

\subsubsection{Architecture}

\begin{figure}[tb]
    \centering
    \includegraphics[width=3in]{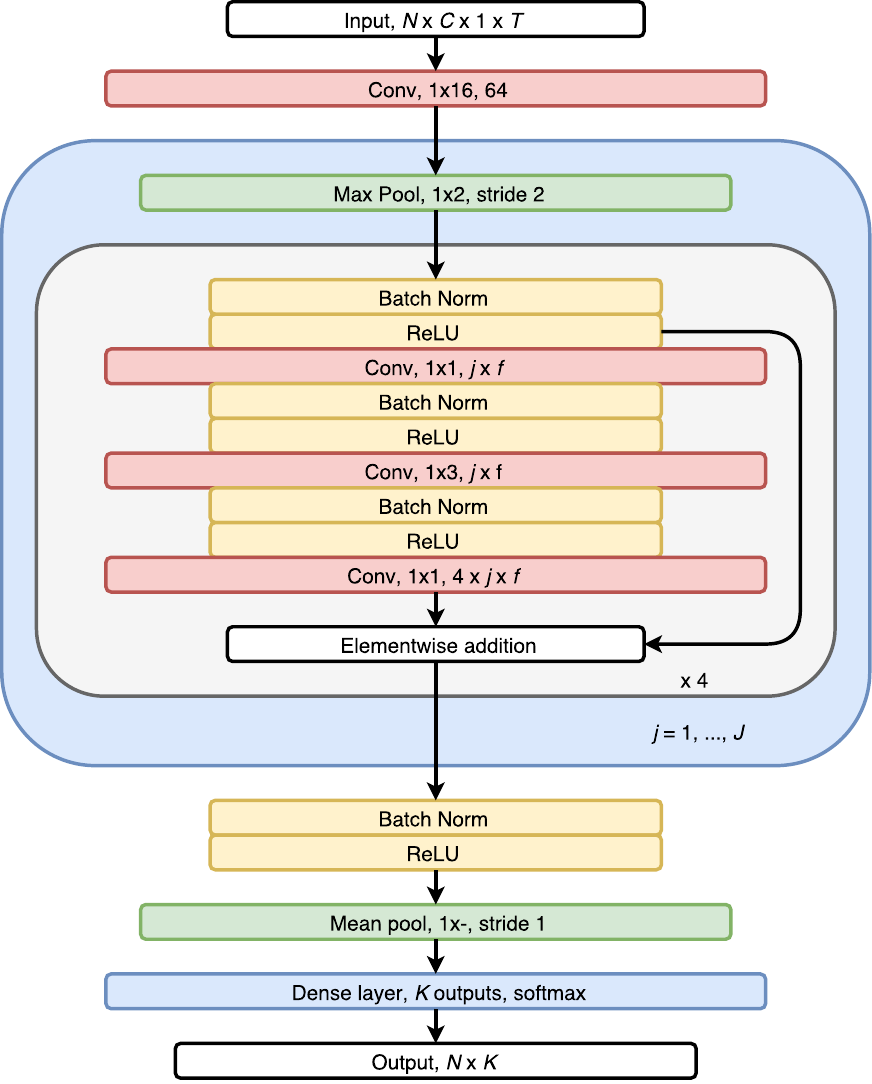}
    \caption{Model architecture. The input tensor has shape $(N, C, 1, T)$, where $N$, $C$, $T$ correspond to the batch size, number of signals, and length of each 30 s epoch, respectively. The output tensor has shape $N\times K$ with $K=5$ sleep stages, while $J=4$, and $f=16$ is the number of block layers and base number of filters.}
    \label{fig:network}
\end{figure}

The residual network model is illustrated in~\cref{fig:network}. Briefly, the bulk network comprised 50 convoluational (conv) and dense layers arranged in four block layers of four bottlenecked residual blocks each. 

A single bottleneck residual block contains three triplets of a batch normalization layer, a rectified linear unit (ReLU) activation layer, and a conv layer. This pre-activation configuration has shown benefits with regards to trainability and generalization compared to vanilla residual blocks~\cite{He2016b}. Projection shortcuts were used between the first ReLU and conv layers to the output of the last conv layer. Kernel sizes were set to $1\times1$ for the first and third conv layers, and $1\times3$ for the second conv layer. The number of output filters for each residual block was $l\times f$ with $l$ being the block layer index and $f=16$, resulting in a total of 256 filters after the final conv layer.

Before the bottleneck blocks, the input tensor $\mathbf{X}$ was passed through an initial conv layer consisting of 64 $1\times16$ filters, and then through a maximum pooling (max pool) layer with a $1\times2$ kernel and stride size, effectively reducing the time-resolution by a factor of 2. This max pool operation was implemented in the beginning of each block layer.

The output tensor from the block layers was subsequently passed to a final batch normalization and ReLU activation layer, followed by a mean pooling layer to reduce the tensor to $\mathbf{X} = \left(x_{nk}\right) \in \mathbb{R}^{N\times 256}$. Finally, a fully connected layer with $K=5$ output units corresponding to the sleep stages resulted in the following output tensor
\begin{equation}
    \mathbf{P} = \left(p_{nk}\right) \in \mathbb{R}^{N \times K}, \quad p_{nk} = \frac{\exp{z_{nk}}}{\sum_{k}^{K} \exp z_{nk}}
\end{equation}
with $p_{nk}$ containing the softmax activations of the output units $z_{nk}$ from the fully connected layer for the $n$th subject and the $k$th sleep stage. The predicted class for the $n$th subject can then be calculated as
\begin{equation}\label{eq:argmax}
    \hat{y}_n = \argmax_{k} p_{nk}.
\end{equation}

\subsubsection{Training}
The optimization problem was constructed using cross entropy loss across $K$ classes and $N$ epochs as objective function, such that
\begin{equation}
    \mathcal{L}(\mathbf{p}_{n} |\, \mathbf{y}_{n},\mathsf{W}) = -\sum_{k=1}^{K}{y_{nk}\log{p_{nk}}},
\end{equation}
is the calculated cross entropy loss for epoch $n$ given predicted class probabilities $\mathbf{p}_n$, true class labels $\mathbf{y}_n$, and the set of current weights $\mathsf{W}$. Then, the average cost across a batch of data is
\begin{equation}
    \mathcal{C}(\mathbf{P} |\, \mathbf{Y},\mathsf{W}) = \frac{1}{N}\sum_{n=1}^{N}{\mathcal{L}(\mathbf{p}_{n} | \mathbf{y}_{n},\mathsf{W})}\label{eq:cost}.
\end{equation}
The cost function was optimized using the Adam optimization algorithm with default hyperparameters~\cite{Kingma2015}. Weights were initialized using variance scaling~\cite{He2015a}, and we applied weight decay during training with $\lambda=10^{-4}$. The initial learning rate was set to $\alpha=10^{-3}$ and was multiplied by $0.1$ every 50000 steps.

In order to investigate the effect of the imbalanced data on the network performance, we trained the following three different configurations. First, we defined a \textit{baseline} configuration as described in the previous sections. The second was a \textit{weighted} configuration, where the cost function in~\cref{eq:cost} was replaced with an average weighted by the inverse frequency for the correct class, such that
\begin{equation}
    \mathcal{C}(\hat{\mathbf{Y}} |\, \mathbf{Y},\mathsf{W}) = \frac{\sum_{n}^{N} \omega_{n}(\mathbf{y}_{n}) \mathcal{L}(\hat{\mathbf{y}}_{n} | \mathbf{y}_{n},\mathsf{W})}{\sum_{n}^{N} \omega_{n}(\mathbf{y}_{n})},
\end{equation}
where $\omega_{n}(\mathbf{y}_{n})$ is the inverse frequency for the correct class for the $n$th subject in the current batch. Finally, a \textit{balanced} configuration was tested, in which we performed resampling of the training dataset in order to balance classes. We oversampled the N1, N3, and REM classes with replacement, while undersampling the N2 class in order to have approximately equal fractions of each class in total.


Models were implemented in TensorFlow 1.4, and trained on a single workstation running Ubuntu 16.04 with a Ryzen 7 1700X 8-core CPU, an NVIDIA GTX 1080 Ti GPU with 11 GB memory, and 32 GB RAM memory.

\subsection{Performance metrics}
Individual precision, recall and F1 scores (Pr, Re, F1) were calculated for each sleep stage and subsequently aggregated for each recording by stage frequency weighting, such that
\begin{equation}
    \mathrm{Pr}_{nk} = \frac{\mathrm{TP}}{\mathrm{TP} + \mathrm{FP}}, \quad \mathrm{Pr}_{n} = \frac{\sum_{k}{\beta_{nk} \mathrm{Pr}_{nk}}}{\sum_{k}\beta_{nk}}
\end{equation}
\begin{equation}
    \mathrm{Re}_{nk} = \frac{\mathrm{TP}}{\mathrm{TP} + \mathrm{FN}}, \quad \mathrm{Re}_{n} = \frac{\sum_{k}{\beta_{nk} \mathrm{Re}_{nk}}}{\sum_{k}\beta_{nk}}
\end{equation}
\begin{equation}
    \mathrm{F1}_{nk} = 2 \cdot \frac{\mathrm{Pr}_{nk} \cdot \mathrm{Re}_{nk}}{\mathrm{Pr}_{nk} + \mathrm{Re}_{nk}}, \quad \mathrm{F1}_{n} = \frac{\sum_{k}{\beta_{nk} \mathrm{F1}_{nk}}}{\sum_{k}\beta_{nk}},
\end{equation}
where $\beta_{nk}$ is the frequency of stage $k$ for recording $n$, and TP, FP and FN are true positives, false positives, and false negatives, respectively. Overall accuracy (Acc) and Cohen's kappa ($\kappa$) were also calculated for each recording. All metrics were summarized by mean and standard deviations.

\subsection{Statistical tests}
Demographic and PSG variables were tested with \textsc{anova}s after establishing normality, while gender was tested with a $\chi^{2}$ test. Significance was set at $p=0.05$.

\section{Results and Discussion}
Performance metrics for the train and eval subgroups are shown in~\cref{tab:train_eval_performance}. Not accounting for Pr, the baseline configuration compares favorably to the weighted and balanced configurations on both subgroups with an average accuracy of \SI{85.0}{\percent} and a Cohen's kappa of \SI{75.4} on the eval subgroup. Since the training data is imbalanced in favor of N2, it would be fair to assume overfitting to the majority class, however, the lower spread in both precision and recall does not support this.
\begin{table}[tb]
    \renewcommand{\arraystretch}{1.3}
    \centering
    \caption{Averaged performance metrics for configurations across train and eval subgroups with best shown in bold.}
    \label{tab:train_eval_performance}
    \begin{tabular}{@{}clccc@{}}
        \toprule
                               &          & baseline          & weighted          & balanced          \\ \midrule
        \multirow{5}{*}{Train} & Acc (\%)      & $ \mathbf{86.1 \pm 5.5} $ & $ 79.4 \pm 7.1 $ & $ 80.4 \pm 7.3 $ \\
                               & $\kappa$ (\%) & $ \mathbf{77.1 \pm 8.6} $ & $ 69.5 \pm 9.7 $ & $ 70.7 \pm 9.8 $ \\
                               & Pr (\%)       & $ 87.1 \pm 4.9 $ & $ 88.7 \pm 4.1 $ & $ \mathbf{88.9 \pm 4.0} $ \\
                               & Re (\%)       & $ \mathbf{86.1 \pm 5.5} $ & $ 79.4 \pm 7.1 $ & $ 80.4 \pm 7.3 $ \\
                               & F1 (\%)       & $ \mathbf{85.3 \pm 6.1} $ & $ 81.8 \pm 6.6 $ & $ 82.6 \pm 6.9 $ \\ \midrule
        \multirow{5}{*}{Eval}  & Acc (\%)      & $ \mathbf{85.0 \pm 6.1} $ & $ 78.4 \pm 7.3 $ & $ 79.7 \pm 7.4 $ \\
                               & $\kappa$ (\%) & $ \mathbf{75.4 \pm 9.5} $ & $ 68.1 \pm 10.5 $ & $ 69.7 \pm 10.0 $ \\
                               & Pr (\%)       & $ 86.3 \pm 5.3 $ & $ 87.8 \pm 4.8 $ & $ \mathbf{88.0 \pm 4.9} $ \\
                               & Re (\%)       & $ \mathbf{85.0 \pm 6.1} $ & $ 78.4 \pm 7.3 $ & $ 79.7 \pm 7.4 $ \\
                               & F1 (\%)       & $ \mathbf{84.0 \pm 7.2} $ & $ 80.7 \pm 7.1 $ & $ 81.9 \pm 7.1 $ \\ \bottomrule
    \end{tabular}
\end{table}
\begin{table}[tb]
    \renewcommand{\arraystretch}{1.3}
    \setlength{\tabcolsep}{4pt}
    \caption{Aggregated confusion matrix and stage-specific performance metrics in test subgroup.}
    \label{tab:test_performance_aggregated}
    \centering
    \begin{tabular}{@{}lcccccccc@{}}
        \toprule
                 & W     & N1   & N2    & N3   & REM   & Pr (\%) & Re (\%) & F1 (\%) \\ \midrule
        W        & 37980 & 1322 & 852   & 2    & 327   & 84.3    & 93.8    & 88.8    \\
        N1       & 3922  & 8784 & 3545  & 0    & 2193  & 51.9    & 47.6    & 49.7    \\
        N2       & 1756  & 5136 & 99564 & 1091 & 991   & 88.6    & 91.7    & 90.2    \\
        N3       & 18    & 1    & 7932  & 4063 & 14    & 78.8    & 33.8    & 47.3    \\
        R        & 1361  & 1680 & 465   & 0    & 23931 & 87.2    & 87.2    & 87.2    \\ \bottomrule
    \end{tabular}
\end{table}
\begin{figure*}[tb]
    \centering
    \includegraphics[width=\textwidth]{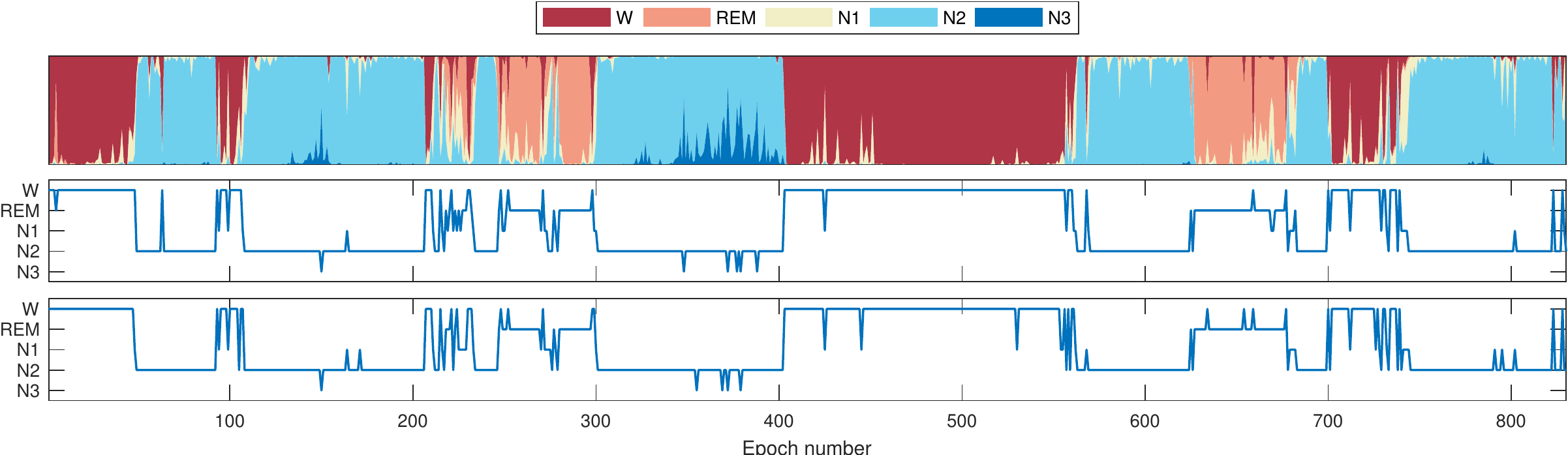}
    \caption{Top: hypnodensity graph of per-epoch probability distributions, middle: automatically scored hypnogram by applying~\cref{eq:argmax}, bottom: manually scored hypnogram. Note the intrusions of N3 into N2 around epoch 150 and 370, and N1 into W around 420.}
    \label{fig:hypnodensity}
\end{figure*}
Evaluating the baseline model on the test subgroup gave only a slight drop in accuracy and $\kappa$, indicating that the model generalizes well, see~\cref{tab:test_performance_aggregated} and~\cref{tab:test_performance_average}. The lowest sensitivity is obtained for N1 and N3, which is in accordance with clinical experience reported in the literature~\cite{Younes2017, Rosenberg2013, Norman2000, Younes2016}. N1 is a transitional stage between wakefulness, drowsiness and sleep often containing beta and alpha activity in epochs of low interscorer agreement, which explains the low predictive power 
in the confusion matrix. The sleep continuum is also apparent in~\cref{fig:hypnodensity} which shows the manually and automatically scored hypnograms in the middle and bottom traces, and the hypnodensity graph in the top trace for a representative subject in the test subgroup. The hypnodensity is a probabilistic representation of the hypnogram, which has found use in the detection of Parkinson's and narcolepsy~\cite{Koch2014, Christensen2014, Stephansen2017}.
\begin{table}[tb]
    \centering
    \caption{Performance across recordings in test subgroup.}
    \label{tab:test_performance_average}
    \begin{tabular}{@{}ccccc@{}}
    \toprule
        Acc              & $\kappa$         & Pr               & Re               & F1               \\ \midrule
        $ 84.1 \pm 6.9 $ & $ 0.746 \pm 0.099 $ & $ 85.7 \pm 6.1 $ & $ 84.1 \pm 6.9 $ & $ 83.1 \pm 7.6 $ \\ \bottomrule
    \end{tabular}
\end{table}
Our baseline model attains favorable performance when comparing to the results reported for the raw waveform CNN model in~\cite{Biswal2017} with both higher accuracy and Cohen's kappa. However, it should be stressed that~\cite{Biswal2017} used EEG from 9000 recordings, while our model uses EEG, EOG and EMG from 1850 recordings. Furthermore, our baseline model performs only slightly worse compared to the best-performing model using manual feature engineering and RNNs in~\cite{Biswal2017}. This indicates a possible performance gain by adding recurrent networks, such as long short-term memory cells, to our network.

A possible limiting factor to our model is the filter kernels. 
The small filter sizes in block layers might not be able to accurately capture the physiological dynamics, but there are indications that many, smaller kernels are preferable to fewer, larger kernels when comparing model complexity versus computational costs~\cite{Szegedy2015}.

Future work will include adding more data to balance classes, and adding long short-term memory cells to the network in order to model temporal dynamics between epochs. As we performed minimal hyperparameter tuning in this work, investigating the effects of changing the network specifications to optimize performance is also a relevant area of future research.

\section{Conclusion}
We have shown that common data transformations such as spectrograms are not necessary for automatic sleep staging. Combining residual learning networks and raw PSG waveforms, we obtained an average accuracy of 84.1\% and Cohen's kappa of 0.745, improving on previously reported results on raw PSG sleep staging. Further testing on independent cohorts will illuminate the clinical applicability of this method, while introducing more data and memory cells will be explored to increase performance even further.




\bibliographystyle{IEEEtran}
\bibliography{IEEEabrv,library.bib}

 \end{document}